\documentclass[10pt,journal,compsoc]{IEEEtran}
\usepackage{subfigure}
\usepackage{graphicx}
\usepackage{multirow}
\usepackage{algorithm}
\usepackage{algorithmic}
\usepackage{url}
\usepackage{balance}
\usepackage{amsmath,amssymb,amsfonts}
\newtheorem{definition}{Definition}

\newtheorem{lemma}{Lemma}

\ifCLASSOPTIONcompsoc
  \usepackage[nocompress]{cite}
\else
  \usepackage{cite}
\fi
\ifCLASSINFOpdf
\else
\fi
\begin{document}
%
\title{DNN2LR: Interpretation-inspired Feature Crossing for Real-world Tabular Data}
%
%
%
%

\author{Zhaocheng Liu,
        Qiang Liu,
        Haoli Zhang,
        Yuntian Chen
}

%
%

\markboth{Journal of \LaTeX\ Class Files,~Vol.~14, No.~8, August~2015}%
{Shell \MakeLowercase{\textit{et al.}}: Bare Demo of IEEEtran.cls for Computer Society Journals}
\IEEEtitleabstractindextext{%
\begin{abstract}
For sake of reliability, it is necessary for models in real-world applications to be both powerful and globally interpretable.
Simple classifiers, e.g., Logistic Regression (LR), are globally interpretable, but not powerful enough to model complex nonlinear interactions among features in tabular data.
Meanwhile, Deep Neural Networks (DNNs) have shown great effectiveness for modeling tabular data, but is not globally interpretable.
In this work, we find local piece-wise interpretations in DNN of a specific feature are usually inconsistent in different samples, which is caused by feature interactions in the hidden layers.
Accordingly, we can design an automatic feature crossing method to find feature interactions in DNN, and use them as cross features in LR.
We give definition of the interpretation inconsistency in DNN, based on which a novel feature crossing method called DNN2LR is proposed.
Extensive experiments have been conducted on four public datasets and two real-world datasets.
The final model, i.e., a LR model empowered with cross features, generated by DNN2LR can outperform the complex DNN model, as well as several state-of-the-art feature crossing methods.
The experimental results strongly verify the effectiveness and efficiency of DNN2LR, especially on real-world datasets with large numbers of feature fields.
\end{abstract}

\begin{IEEEkeywords}
Automated Machine Learning, Feature Crossing, Deep Neural Networks, Interpretability, Logistic Regression.
\end{IEEEkeywords}}

\maketitle

\IEEEdisplaynontitleabstractindextext

%
\IEEEpeerreviewmaketitle

\IEEEraisesectionheading{\section{Introduction}\label{sec:introduction}}

%
%
%
%
\IEEEPARstart{I}{n} application areas such as finance and healthcare, reliability and interpretability are strongly desired.
Thus, powerful and globally interpretable models are well appreciated in real-world applications.
In \cite{guidotti2018survey}, the global interpretability is defined as, \emph{we are able to understand the whole logic of a model and follow the entire reasoning leading to all the different possible outcomes}.
Some commonly-used classifiers, e.g., Logistic Regression (LR), are simple and globally interpretable.
However, LR usually has relatively poor performances, and it is hard for LR to model complex nonlinear interactions among features in tabular data.
To improve the performances of LR, heavy handcrafted feature engineering is usually required.
On the other hand, Deep Neural Networks (DNNs) \cite{zhang2016deep} and tree ensemble models \cite{chen2016xgboost,ke2017lightgbm} are able to take use of complex nonlinear feature interactions.
However, as pointed in \cite{guidotti2018survey,hall2019introduction}, these methods are not globally interpretable.
That is to say, we need to trade-off between the performance and the interpretability \cite{hall2019introduction}.
Fortunately, automatic feature crossing, which takes cross-product of categorical features, is a practical direction in automated machine learning \cite{quanming2018taking}, as well as a key direction for automatic feature engineering.
It is a promising way to capture the interactions among categorical features in tabular data in real-world applications \cite{chapelle2015simple,yuanfei2019autocross}.
Via automatic generation of cross features, we can achieve better performances with the simple LR model without heavy handcrafted feature engineering.
As pointed in \cite{yuanfei2019autocross}, cross features, instead of latent embeddings or latent representations in DNN, are highly interpretable.
Moreover, utilizing cross features is more flexible and suitable for large-scale online tasks \cite{cheng2016wide,yuanfei2019autocross}.
With the research on automatic feature crossing, we can make the simple LR model powerful and globally interpretable simultaneously, and thus reliable models can be obtained in real-world applications.

Previous works on feature crossing mostly try to search in the set of possible cross feature fields \cite{rosales2012post,lou2013accurate,chapelle2015simple,katz2016explorekit,khurana2018feature,yuanfei2019autocross}. The candidate set for searching is usually inevitably large, which leads to low efficiency of learning feature crossing.
Tree ensemble models \cite{chen2016xgboost} have also been utilized for finding cross features \cite{he2014practical}. However, as pointed in \cite{ke2019deepgbm}, it is hard for these tree ensemble models to well handle sparse categorical features, which are important in real-world applications \cite{liu2002discretization,chapelle2015simple}.
Besides above crossing methods, some factorization-based methods \cite{rendle2010factorization,liu2015cot,blondel2016higher,liu2020autofis} seek to model feature interactions with product-based similarity. However, the product-based similarity is hard to model all kinds of feature interactions in various applications.
Meanwhile, some deep learning-based CTR prediction methods design various crossing modules for modeling feature interactions \cite{qu2016product,cheng2016wide,wang2017deep,guo2017deepfm,lian2018xdeepfm,song2018autoint}. Unfortunately, due to the hidden projections in the crossing modules, which are usually incorporated in deep learning technologies, most of these deep learning-based methods are not globally interpretable and not capable to explicitly generate all the cross features captured in the models.

As mentioned in previous works \cite{zhang2016deep,qu2016product,guo2017deepfm}, DNN can be a powerful method for capturing various feature interactions in its hidden layers.
DNN can implicitly interact features, but can not explicitly provide interpretable cross features.
Recently, the interpretability of deep models has drawn great attention in academia, and research works mostly focus on piece-wise interpretability, which means assigning a piece of local interpretation for each sample \cite{simonyan2013deep,selvaraju2017grad,alvarez2018towards,li2015visualizing,smilkov2017smoothgrad,yuan2019interpreting}.
That is to say, a nonlinear DNN model can be regarded as a combination of numbers of linear classifiers \cite{montufar2014number,ribeiro2016should,chu2018exact,cong2020exact}.
This process can be done via the gradient backpropagation from the prediction layer to the input features.
We observe that, local interpretations of a specific feature are usually inconsistent in different samples.
In this work, we show that, such inconsistency is caused by feature interactions occurred in the hidden layers of DNN.
Accordingly, we give definition of the interpretation inconsistency in DNN, which can help us find useful cross features.
With the interpretation inconsistency in DNN, we can take advantage of the strong expressive ability of DNN and the good interpretability of LR at the same time.

\begin{figure}
\centering
\includegraphics[width=0.45\textwidth]{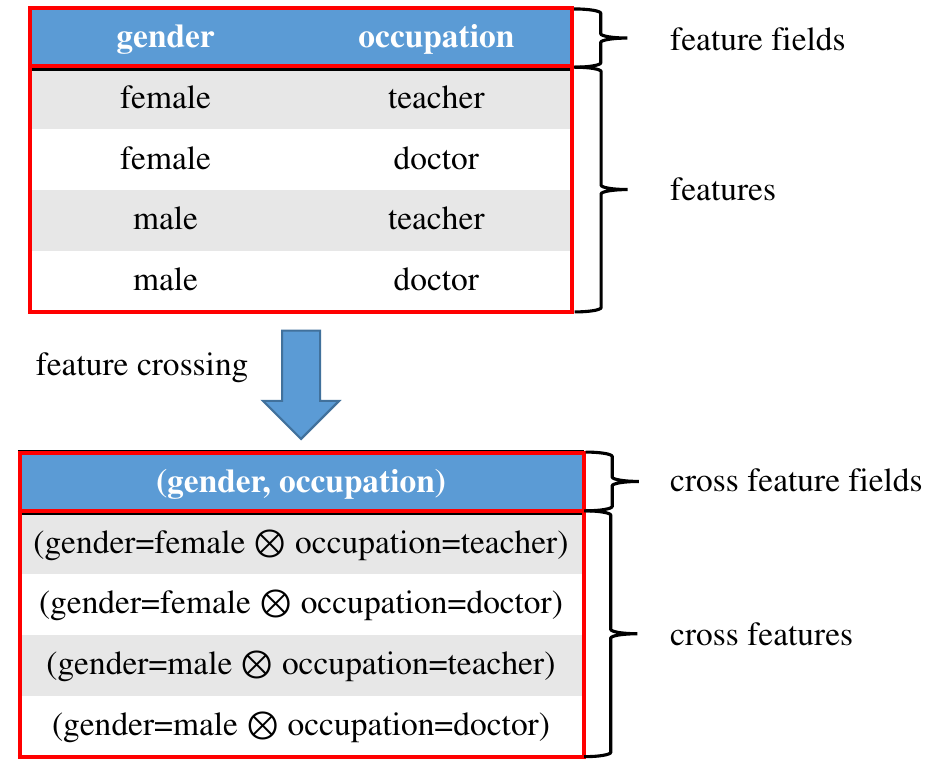}
\caption{An example of feature crossing.}
\label{fig:example}
\end{figure}

We propose a novel method called DNN2LR, which can automatically learn useful cross feature fields from the interpretation inconsistency of DNN.
The process of DNN2LR can be detailed as follows:
(1) We train a DNN model with the whole training set.
(2) For each sample in the validation set, we calculate the interpretation inconsistency value of each feature. If the value of interpretation inconsistency is larger than a threshold, we use the corresponding feature field for conducting feature crossing.
(3) Then, we generate a global candidate set containing both second-order and higher-order cross feature fields.
(4) Based on the candidate set, we can train a simple LR model, therefore rank and select useful feature fields according to their contribution measured on the validation set.
(5) Finally, we can obtain the set of useful cross feature fields, as well as a LR model empowered with cross features.

We conduct experiments on four public datasets, as well as two real-world datasets.
According to the experimental results, a simple LR model empowered with the final set of cross feature fields achieves better performances comparing with the complex DNN model, as well as some state-of-the-art feature crossing methods, e.g., AutoCross \cite{yuanfei2019autocross} and AutoFM \cite{liu2020autofis}.
To achieve powerful performances, a candidate set with only $2N$ or $3N$ cross feature fields is enough, where $N$ is the number of original feature fields in each dataset.
To be noted, this is extremely small compared to the whole set of second-order and higher-order cross feature fields, especially when $N$ is large.
With such a compact candidate set, the searching for final useful cross feature fields can be efficiently done.
According to our experiments, on wide tabular datasets with large numbers of feature fields, compared with newly-proposed state-of-the-art feature crossing methods AutoCross and AutoFM, DNN2LR can reduce the running time by $6\times$ to $50\times$.
Meanwhile, as the final model is a LR model, it is globally interpretable.
In a word, with the proposed DNN2LR method, we can obtain powerful and globally interpretable models in an efficient way.

The main contributions of this paper are summarized as follows:
\begin{itemize}
\item We give definition of the interpretation inconsistency in DNN, and accordingly propose a novel DNN2LR method. The DNN2LR method can generate an accurate candidate set of cross feature fields, with relatively small amount compared to the whole set of second-order and higher-order cross feature fields.
\item Useful cross feature fields can be directly ranked and selected based on the candidate set and corresponding contribution in a LR model. The whole process of learning feature crossing can be done via simply training a DNN model and a LR model.
\item Extensive experiments have been conducted on several datasets. The final model, i.e., a LR model empowered with cross features, generated by DNN2LR can outperform the complex DNN model, as well as the state-of-the-art feature crossing methods. The high efficiency of DNN2LR is also strongly verified, especially on real-world datasets with large numbers of feature fields.
\end{itemize}

The rest of the paper is organized as follows.
In section \ref{sec:related}, we first summarize some works on feature crossing and interpretability.
Section \ref{sec:interpretation} gives brief analysis about why and how we learn feature crossing from interpretations of DNN.
Section \ref{section:DNN2LR} details our proposed DNN2LR method.
In section \ref{section:exp}, we report and discuss our experimental results.
And finally, section \ref{sec:conclusion} concludes our work.

\section{Related Work} \label{sec:related}

In this section, we review some works on feature crossing and interpretability of DNN.

\subsection{Feature Crossing} \label{sec:related:FC}

According to the definition in previous works \cite{cheng2016wide,yuanfei2019autocross}, in a specific sample $k$, we can conduct $n$th-order feature crossing as
\begin{equation} \label{equation:crossing}
\mathop g\nolimits_{\mathop x\nolimits_{\mathop k, f_1} ,\mathop x\nolimits_{\mathop k, f_2} ,...,\mathop x\nolimits_{\mathop k, f_n} }  = \mathop x\nolimits_{\mathop k, f_1} \otimes \mathop x\nolimits_{\mathop k, f_2} \otimes ... \otimes \mathop x\nolimits_{\mathop k, f_n},
\end{equation}
where $\otimes$ denotes cross-product, $\mathop g\nolimits_{\mathop x\nolimits_{\mathop k, f_1} ,\mathop x\nolimits_{\mathop k, f_2} ,...,\mathop x\nolimits_{\mathop k, f_n} }$ is the corresponding generated cross feature, and $\mathop x\nolimits_{\mathop k, f}$ is a binary feature associated with categorical feature field $\mathop f$, e.g., feature ``occupation=teacher'' associated with the field ``occupation''. Via feature crossing, the cross feature of a female teacher can be denoted as (``gender=female'' $\otimes$ ``occupation=teacher''). And an example of feature crossing can be found in Fig. \ref{fig:example}.
Moreover, considering feature discretization has been proven useful to improve the capability of numerical features \cite{liu2002discretization,chapelle2015simple,franc2018learning,yuanfei2019autocross,liu2020an}, we can conduct feature discretization on numerical feature fields to generate corresponding categorical feature fields. Thus, we are able to perform feature crossing on both categorical and numerical feature fields.

It is a direct way to search for useful cross feature fields in a candidate set.
However, the candidate set for searching is usually inevitably large, which leads to low efficiency of learning feature crossing.
Most searching-based methods focus on generating second-order features
\cite{rosales2012post,lou2013accurate,chapelle2015simple,kanter2015deep,katz2016explorekit,nargesian2017learning,kaul2017autolearn}.
In \cite{chapelle2015simple}, the authors try to generate and select second-order cross feature fields according to Conditional Mutual Information (CMI). However, once the mutual information of an original feature field is high, the generated cross feature fields containing it will also have high conditional mutual information.
AutoLearn \cite{kaul2017autolearn} selects cross feature fields by using regularized regression models, where it is hard to learn a regression model for all the cross feature fields on a wide dataset.
Some works \cite{katz2016explorekit,nargesian2017learning} takes meta-learning into consideration for feature generation. However, the effectiveness of meta-learning in these methods remains a question and requires extremely large amount of data.
There are also some methods incorporating genetic algorithm \cite{tran2016genetic} and reinforcement learning \cite{khurana2018feature,chen2019neural} for finding feature combinations. However, with genetic algorithm or reinforcement learning, we still have a large space to explore.
Moreover, as introduced in \cite{khurana2018feature}, it also requires large amount of data for the training of reinforcement learning.
To tackle with above problems, AutoCross \cite{yuanfei2019autocross} presents a framework to search in the large candidate set more efficiently, which is a greedy and approximate alternative: (1) AutoCross iteratively searches in a set of cross feature fields, where the set is initialized as all the second-order feature fields, and the selected cross feature field is greedily used to generate new high-order cross feature fields in the next iteration. (2) AutoCross uniformly divides the dataset into at least $\sum\nolimits_{i = 0}^{\left\lceil {\mathop {\log }\nolimits_2 t} \right\rceil  - 1} {\mathop 2\nolimits^i }$ batches, where $t$ is the size of candidate set in AutoCross, and iteratively train a field-wise LR model on part of the data to validate the contribution of a cross feature field.
The authors apply the generated cross features, containing both second- and high-order cross features, in a LR model, and it achieves approximate or even better performances comparing with the complex DNN model on $10$ datasets.
However, AutoCross still searches for useful cross feature fields in a large candidate set, whose size shows exponential relation with the number of the original feature fields. For example, when the size of original feature fields is $10$, the number of second-order cross feature fields is $45$. And when the size of original feature fields becomes $100$, $200$, $500$ and $1000$, the number of second-order cross feature fields becomes $4950$, $19900$, $124750$ and $499500$ respectively. When the candidate set is large, the searching efficiency will still be low.
Moreover, when the candidate set is large, data for training the field-wise LR model of a candidate cross feature field will be too little to produce reliable results. Therefore, the results of AutoCross are with some randomness, especially on some wide tabular datasets.

Some works \cite{he2014practical,shi2020safe} utilize tree ensemble model, e.g., GBDT (Gradient Boosting Decision Tree) \cite{ke2017lightgbm} and XGBoost \cite{chen2016xgboost}, for generating cross features.
In \cite{he2014practical}, each tree in the GBDT model corresponds to a cross feature field, and the leaf node in a tree corresponds to a cross feature.
In \cite{shi2020safe}, the authors take use of split features in XGBoost for generating cross features.
As pointed in \cite{ke2019deepgbm}, it is hard for tree ensemble models to well handle sparse discrete features, which are essential in real-world applications.
This constrains the effectiveness and application scenarios of tree-based feature crossing methods.

\begin{table*}[htbp]
\centering
\caption{The values of interpretation inconsistency in DNN of different features associated with two feature fields, where $\alpha \in \{0,1\}$ and $\beta \in \{0,1\}$, on four toy datasets, characterizing logical operations AND, OR, XNOR and XOR respectively.}\label{tab:toy_data}
    \begin{tabular}{|cc|cc|cc|cc|cc|}
\hline
    \multicolumn{2}{|c|}{sample} & \multicolumn{2}{c|}{AND} & \multicolumn{2}{c|}{OR} & \multicolumn{2}{c|}{XNOR} & \multicolumn{2}{c|}{XOR} \\
\hline
    $\alpha$     & $\beta$     & $\alpha$     & $\beta$     & $\alpha$     & $\beta$     & $\alpha$     & $\beta$     & $\alpha$     & $\beta$ \\
\hline
    0     & 0     & 0.0000  & 0.0000  & 0.0006  & 0.0002  & 0.0322  & 0.0155  & 0.0049  & 0.0042  \\
    0     & 1     & 0.0000  & 0.0001  & 0.0001  & 0.0008  & 0.0030  & 0.0046  & 0.0007  & 0.0005  \\
    1     & 0     & 0.0001  & 0.0000  & 0.0000  & 0.0003  & 0.0021  & 0.0023  & 0.0002  & 0.0002  \\
    1     & 1     & 0.0001  & 0.0000  & 0.0001  & 0.0001  & 0.0139  & 0.0163  & 0.0026  & 0.0056  \\
\hline
    \end{tabular}%
\end{table*}%

As an extended method of Matrix Factorization (MF), Factorization Machine (FM) \cite{rendle2010factorization,rendle2012factorization} has been a successful way to capture second-order feature interactions.
Field-aware FM (FFM) \cite{juan2016field} incorporates field-aware interactions between different feature fields.
To overcome the problem that conventional FM can only model second-order feature interactions, Higher-Order FM (HOFM) \cite{blondel2016higher} proposes to model higher-order feature interactions in the FM architecture, and a linear-time algorithm is presented.
Meanwhile, Interaction Machine (IM) \cite{yu2020deep} proposes to capture higher-order FM based on Newton's identities.
Besides, FM has recently been extended via combining with DNN architectures, e.g., Neural Factorization Machines (NFM) \cite{he2017neural}, DeepFM \cite{guo2017deepfm} and DeepIM \cite{yu2020deep}.
The calculation of feature interactions in FM is somehow product-based similarity.
This may be suitable for some kinds of feature interactions, e.g., the matching and correlating in the scenario of recommendation \cite{rendle2011fast}.
However, it is hard for these factorization-based methods to capture all kinds of feature interactions in various real-world application scenarios.
Another drawback of FM is that, it models all feature interactions of specific-order, most of which are not useful for making predictions.
To deal with this problem, Automatic Feature Interaction Selection (AutoFIS) \cite{liu2020autofis} directly learns the weight for each cross feature field, for both second-order and higher-order, and thus an AutoFM model is obtained.
It promotes the performances of FM, but faces the problem of high computational cost, especially when the dataset is wide and the desired interaction order is high.

Nowadays, deep learning-based prediction methods have shown their effectiveness.
Among these methods, some try to generate and represent cross features via designing various deep learning-based crossing modules or crossing layers.
These methods mainly focus on the task of CTR prediction.
For better performances, these crossing modules are usually applied along with DNN architectures.
The Wide \& Deep model \cite{cheng2016wide} directly learns parameters of manually designed cross features in the wide component.
The Product-based Neural Network (PNN) \cite{qu2016product} applies inner-product or outer-product to capture second-order features.
In Deep \& Cross Network (DCN) \cite{wang2017deep}, the authors design an incremental crossing module, named CrossNet, to capture second-order as well as higher-order features.
AutoINT \cite{song2018autoint} is a combination of residual connections \cite{he2016deep} and a crossing module based on Self-Attention \cite{vaswani2017attention}. With the residual connections, the DNN architecture is performed with the input of original features.
In xDeepFM \cite{lian2018xdeepfm}, Compressed Interaction Network (CIN) is proposed as a crossing module based on outer-product calculation of features, and it is performed together with DNN.
Meanwhile, convolutional neural network is also incorporated for modeling local interactions among nearby features \cite{liu2015convolutional,chan2018convolutional,liu2019feature}.
Inherit from deep learning technologies, above feature crossing modules are mostly associated with hidden projection, in each layer or between adjacent layers.
This happens in most deep learning-based methods, including several state-of-the-art crossing modules, e.g., CrossNet \cite{wang2017deep}, CIN \cite{lian2018xdeepfm} and Self-Attention \cite{song2018autoint}.
That is to say, most of deep learning-based methods are not globally interpretable, and are hard to explicitly generate all the useful cross features captured in the models.

\subsection{Interpretability of DNN}

Recently, the interpretability of DNN has drawn great attention in academia, and research works mostly focus on local piece-wise interpretability, which means assigning a piece of local interpretation for each sample \cite{guidotti2018survey}.
Some unified approaches are proposed to fit a linear classifier in each local space of input samples \cite{ribeiro2016should,lundberg2017unified}.
Some works investigate the gradients from the final predictions to the input features in deep models, which can be applied in the visualization of deep vision models \cite{zhou2016learning,selvaraju2017grad,smilkov2017smoothgrad,alvarez2018towards}, as well as the interpretation of language models \cite{li2015visualizing,yuan2019interpreting}.
Perturbation on input features is also utilized to find local interpretations of both vision models \cite{fong2017interpretable} and language models \cite{guan2019towards}.
Meanwhile, via adversarial diagnosis of neural networks, adversarial examples can also be introduced for local interpretation of DNN \cite{koh2017understanding,dong2018towards}.
In some views, attention in deep models can also be regarded as local interpretations  \cite{wang2019towards,sun2020understanding,mohankumar2020towards}.
As discussed in some works \cite{ribeiro2016should,lundberg2017unified}, the nonlinear DNN model can be regarded as a combination of numbers of linear classifiers, and the upper bound of the number in DNN with piece-wise linear activations has been given \cite{montufar2014number}.
Moreover, piece-wise linear DNN has been exactly and consistently interpreted as a set of linear classifiers \cite{chu2018exact,cong2020exact}.

\begin{figure*}
\centering
\includegraphics[width=0.9\textwidth]{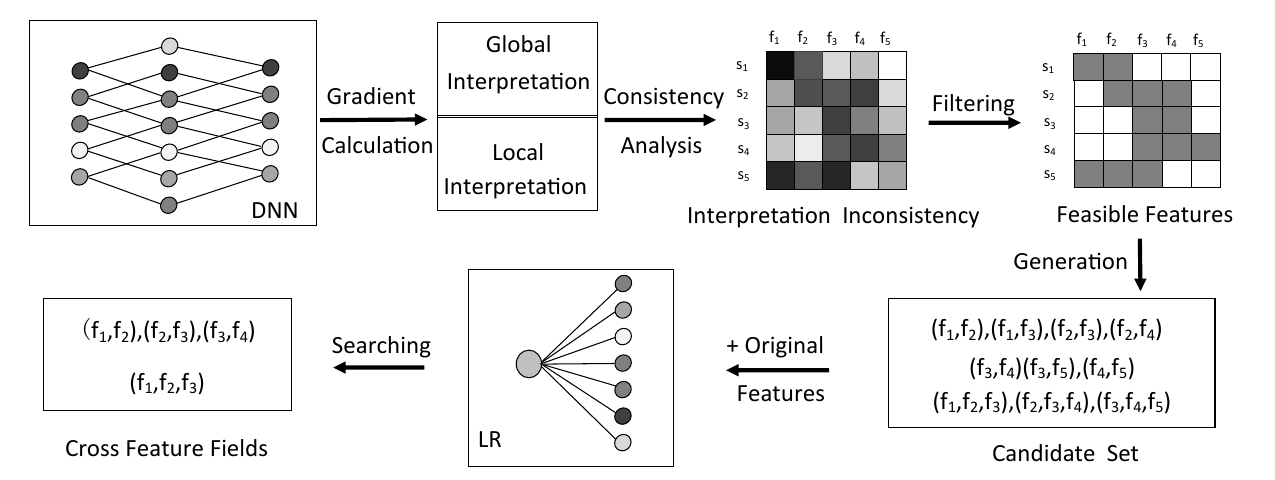}
\caption{Overview of our proposed DNN2LR approach, where $\mathop s\nolimits_i$ and $\mathop f\nolimits_j$ indicate the i-th sample and the j-th original feature field respectively.}
\label{fig:dnn2lr}
\end{figure*}

\section{Learning from Interpretations in DNN} \label{sec:interpretation}

In this section, we present that we can learn feature crossing from interpretations in DNN.

\subsection{Local Interpretations in DNN}

The widely-used DNN model has shown to be capable of capturing various feature interactions in its hidden layers \cite{zhang2016deep,qu2016product,guo2017deepfm}.
This means, if we can explicitly find the feature interactions in DNN and use them as cross features in LR, we can take advantage of the strong expressive ability of DNN and the global interpretability of LR.
Recently, extensive works have been done to study local piece-wise interpretability of DNN, which means a DNN model can be regarded as a combination of numbers of linear classifiers \cite{montufar2014number,ribeiro2016should,chu2018exact,cong2020exact}.
And this process can be done via the gradient backpropagation from the prediction layer to the input features \cite{simonyan2013deep,selvaraju2017grad,alvarez2018towards,li2015visualizing,smilkov2017smoothgrad,yuan2019interpreting}.
Therefore, we can first define the local interpretation in DNN.

\begin{definition}(Local Interpretation) \label{def:local_interpretation}
Given a specific feature $\mathop x\nolimits_{k,f}$ associated with the feature field $\mathop f$ in a specific sample $k$, the corresponding local interpretation is
\begin{equation} \label{equation:local_interpretation}
\mathop I^l\nolimits_{k,f}  = \mathop w\nolimits_{k,f} \mathop e\nolimits_{k,f}^\top,
\end{equation}
where $\mathop e\nolimits_{\mathop k,f}$ denotes the corresponding feature embeddings of $\mathop x\nolimits_{k,f}$ in the DNN model, and $\mathop w\nolimits_{ k,f}$ is the local weights computed via gradient backpropagation
\begin{equation} \label{equation:local_weights}
\mathop w\nolimits_{k,f}  = \frac{{\partial \mathop {\hat y}\nolimits_k }}{{\partial \mathop e\nolimits_{k,f} }},
\end{equation}
where ${\mathop {\hat y}\nolimits_k }$ denotes the prediction made by the DNN model for the sample $k$.
\end{definition}

\subsection{Interpretation Inconsistency}

Local interpretation refers to the contribution of the corresponding feature to the final prediction in the corresponding sample.
Sometimes, local interpretations of a specific feature are inconsistent in different samples.
When local interpretations are consistent in different samples, it means the corresponding feature contributes to the final prediction on its own.
When local interpretations are inconsistent, it means the contribution of the corresponding feature is affected by other features.
The inconsistency of local interpretations is caused by feature interactions in the hidden layers of DNN, and we have the following conclusion.

\begin{lemma} \label{lemma:inconsistency}
If a feature field nonlinearly interacts with other feature fields in the hidden layers of DNN, features associated with the feature field will have inconsistent local interpretations among different samples.
\end{lemma}

Suppose feature fields $\mathop f\nolimits_1 ,\mathop f\nolimits_2 ,......$ are interacted in DNN, i.e., ${{\hat y}_k} = P\left( {{\rm{ }}{e_{k,{\rm{ }}{f_1}}},{\rm{ }}{e_{k,{\rm{ }}{f_2}}},......} \right)$, where $P\left(  \cdot  \right)$ is a nonlinear interaction function.
For example, $\mathop {\hat y}\nolimits_k = \mathop e\nolimits_{k,\mathop f\nolimits_1 } \mathop e\nolimits_{k,\mathop f\nolimits_2 }^ \top$.
Then, we can have the following local interpretation
\begin{equation}
{\rm{ }}I_{k,{\rm{ }}{f_i}}^l = \frac{{\partial P\left( {{\rm{ }}{e_{k,{\rm{ }}{f_1}}},{\rm{ }}{e_{k,{\rm{ }}{f_2}}},......} \right)}}{{{\rm{ }}\partial {e_{k,{\rm{ }}{f_i}}}}}{\rm{ }}e_{k,{\rm{ }}{f_i}}^ \top.
\end{equation}
In the example of $\mathop {\hat y}\nolimits_k = \mathop e\nolimits_{k,\mathop f\nolimits_1 } \mathop e\nolimits_{k,\mathop f\nolimits_2 }^ \top$, the local interpretations are $\mathop I\nolimits_{k,\mathop f\nolimits_1 }^l  = \mathop e\nolimits_{k,\mathop f\nolimits_2 } \mathop e\nolimits_{k,\mathop f\nolimits_1 }^ \top$ and $\mathop I\nolimits_{k,\mathop f\nolimits_2 }^l  = \mathop e\nolimits_{k,\mathop f\nolimits_1 } \mathop e\nolimits_{k,\mathop f\nolimits_2 }^ \top$.
Obviously, the local interpretation of ${\mathop x\nolimits_{k,\mathop f\nolimits_i } }$ is affected by the features associated with other feature fields.
As the values of ${x_{k,{\rm{ }}{f_{i'}}}}(i' \ne i)$ change among different samples, the local interpretations of ${\mathop x\nolimits_{k,\mathop f\nolimits_i } }$ are inconsistent.
In contrast, if there is no nonlinear interaction among $\mathop f\nolimits_1 ,\mathop f\nolimits_2 ,......$, we have an addition form, i.e., ${{\hat y}_k} = {\rm{ }}{P_1}\left( {{\rm{ }}{e_{k,{\rm{ }}{f_1}}}} \right) + {\rm{ }}{P_2}\left( {{\rm{ }}{e_{k,{\rm{ }}{f_2}}}} \right) + ......$, where $\mathop P\nolimits_i \left(  \cdot  \right)$ is an arbitrary function.
Then, we can have the following local interpretation
\begin{equation}
{\rm{ }}I_{k,{\rm{ }}{f_i}}^l = \frac{{\partial {P_i}\left( {{\rm{ }}{e_{k,{\rm{ }}{f_i}}}} \right)}}{{{\rm{ }}\partial {e_{k,{\rm{ }}{f_i}}}}}{\rm{ }}e_{k,{\rm{ }}{f_i}}^ \top.
\end{equation}
Obviously, with the same feature ${{x_{k,{\rm{ }}{f_i}}}}$, there will be consistent local interpretations among different samples.

To measure the degree of inconsistency among local interpretations of a specific feature in different samples, we first need to calculate its global interpretation, and then calculate the interpretation inconsistency between the local interpretation and the global interpretation.

\begin{definition}(Global Interpretation) \label{def:global_interpretation}
Given the feature $\mathop x\nolimits_{\mathop k,f}$, the corresponding global interpretation is
\begin{equation} \label{equation:global_interpretation}
\mathop I^g\nolimits_{k,f}  = \mathop {\bar w}\nolimits_{k,f} \mathop e\nolimits_{k,f}^\top,
\end{equation}
where ${\mathop {\bar w}\nolimits_{\mathop f } }$ is the average local weights of the feature $\mathop x\nolimits_{\mathop k,f}$ in all samples, named as global weights and formulated as
\begin{equation} \label{equation:global_weights}
\small
{{\bar w}_{k,f}} = \frac{1}{{\left\| {\left\{ {\mathop x\nolimits_{k',f}  = \mathop x\nolimits_{k,f} |k' \in \Omega } \right\}} \right\|}}\sum\limits_{k' \in \Omega ,\mathop x\nolimits_{k',f}  = \mathop x\nolimits_{k,f} } {{\rm{ }}{w_{k',f}}},
\end{equation}
where $\Omega$ is the set of samples.
\end{definition}

\begin{definition}(Interpretation Inconsistency) \label{def:interpretation_inconsistency}
Given a specific feature $\mathop x\nolimits_{\mathop k,f}$ in a specific sample $k$, the corresponding interpretation inconsistency is
\begin{equation} \label{equation:interpretation_inconsistency}
{\rm{ }}{d_{k,f}} = \mathop {\left\| {\left( {{\rm{ }}{w_{k,f}} - {{\bar w}_{k,f}}} \right){\rm{ }}e_{k,f}^ \top } \right\|}\nolimits_{\rm{2}}^{\rm{2}}.
\end{equation}
\end{definition}

For the calculation of interpretation inconsistency, we adopt absolute difference, instead of relative difference.
This is because, the values of interpretations indicate the contribution to the final predictions, and features with little contribution are hard to produce useful feature interactions.

\subsection{Empirical Experiments on Toy Datasets}

According to Lemma \ref{lemma:inconsistency}, the interpretation inconsistency in DNN is able to lead us to generate a compact and accurate candidate set of cross feature fields.
And larger the values of interpretation inconsistency of a specific feature, more the corresponding feature field can work for feature crossing.

To verify above conclusion, we conduct empirical experiments on four toy datasets.
The four datasets characterize four different logical operations: AND, OR, XNOR and XOR. And we have two input feature fields, where $\alpha \in \{0,1\}$ and $\beta \in \{0,1\}$. Thus, for each toy dataset, we have four different samples, and the corresponding labels are in $\{0,1\}$.
As we know, the logical operations AND and OR are easy and linearly separable, therefore no cross features are needed. In contrast, the logical operations XNOR and XOR are not linearly separable, therefore second-order cross feature field consisting of $\alpha$ and $\beta$ should be generated.

We train a DNN model on each of the four toy datasets until convergence, where the Area Under Curve (AUC) evaluation becomes $1.0$.
On all datasets, the gradients from the prediction layer to the feature layer is computed, and the interpretation inconsistency is obtained, as shown in Tab. \ref{tab:toy_data}.
It is clear that, interpretation inconsistency values on AND and OR are extremely small, while those on XNOR and XOR are relatively large.
This means, $\alpha$ and $\beta$ should be crossed on XNOR and XOR, while should not on AND and OR.
Thus, the experimental results suggest that, it is proper to learn feature crossing from the interpretation inconsistency in DNN.


\section{DNN2LR}\label{section:DNN2LR}

In this section, we formally propose the DNN2LR approach.
In general, DNN2LR consists of two steps: (1) generating a compact and accurate candidate set of cross feature fields; (2) searching in the candidate set for the final cross feature fields.
Specifically, we use $c_i = \left(\mathop f\nolimits_1, \mathop f\nolimits_2, ..., \mathop f\nolimits_n\right)$ to denote a specific cross feature field generated by crossing original fields $\mathop f\nolimits_1, \mathop f\nolimits_2, ..., \mathop f\nolimits_n$.
Fig. \ref{fig:dnn2lr} provides an overview of the proposed DNN2LR approach.

\subsection{Candidate Set Generation} \label{sec:candidate}

As we rely on the local piece-wise interpretation of DNN to generate the compact and accurate candidate set of cross features, we first need to train a DNN model.
The input of DNN is the original features, which are sparse high dimensional vectors.
Thus, we use an embedding layer \cite{zhang2016deep} to transform the input features into low dimensional dense representations.
Then, the dense representations are passed through some linear transformation and nonlinear activation to obtain the predictions of samples, where we use ReLu as the activation for hidden layers, and sigmoid for the output layer to support binary classification tasks.

Based on the trained DNN model, in each validation sample $k$, we compute the interpretation inconsistency $\mathop d\nolimits_{k,f}$ of the feature $\mathop x\nolimits_{\mathop k ,f}$, as defined in Def. \ref{def:interpretation_inconsistency}.
Therefore, we obtain an interpretation inconsistency matrix $D$, where $\mathop D \left[ k,f \right]$ is the interpretation inconsistency value of the $f$-th feature field in the $k$-th sample.
Then, we conduct an element-wise filtering on matrix $D$ with a threshold $\eta$, which can be formulated as
\begin{equation}
    \mathop D \nolimits^{\rm{*}} \left[ k, f \right] =
    \begin{cases}
        1,& \mathop D \left[ k, f \right] \geq Quantile \left( D, 1 - \eta \right) \\
        0,& otherwise
    \end{cases}
    ,
\end{equation}
where $\mathop D^*$ is a binary feasible feature matrix, and $\mathop D \nolimits^{\rm{*}} \left[ k, f \right]$ indicates whether the $f$-th feature in the $k$-th sample has interacted with other features in the hidden layers.
$Quantile \left( D, 1 - \eta \right)$ denotes the $1 - \eta$ quantile of the matrix $D$, which means keeping the top $\eta$ elements ($0\% < \eta < 100\%$) in $D$ with largest values of interpretation inconsistency.
Then, for each feasible feature $\mathop D \nolimits^{\rm{*}} \left[ k, f \right] = 1$, the corresponding feature field $f$ can be used to generate candidate cross feature fields.

Finally, we greedily generate the candidate set of cross features fields.
Considering that extremely high-order cross features are rarely useful, we construct our candidate set with only 2nd-order, 3rd-order and 4th-order cross feature fields.
To construct a compact and accurate candidate set, we need to find cross feature fields which are most frequently interacted in DNN.
According to the feasible feature matrix $\mathop D^*$, we count the occurrences of possible cross feature fields.
We then rank the cross feature fields in a descending order according to the corresponding occurrence frequency, and select the top $\varepsilon$ cross feature fields as our candidate set.
According to our experiments, to make LR achieve better performances than the complex DNN model, only $\varepsilon = 2N$ or $\varepsilon = 3N$ is needed, where $N$ is the size of the original feature fields.
To be noted, in general, this is extremely small comparing with the size of the entire set of 2nd-order, 3rd-order and 4th-order cross feature fields.
For example, when $N = 100$, the size of the corresponding entire set of cross feature fields will be $4,087,875$ ($4,950 + 161,700 + 3,921,225$).
Accordingly, a compact candidate set of cross feature fields is generated, and efficiently searching for useful cross feature fields can be conducted.

\subsection{Searching for Final Cross Feature Fields}

\begin{algorithm}[t]
    \caption{Searching for Final Cross Feature Fields.}
    \label{alg:searching}
    \begin{algorithmic}[1]
        \REQUIRE candidate set $\mathop S = \{\mathop c\nolimits_1, \mathop c\nolimits_2, ..., \mathop c\nolimits_{\varepsilon}\}$, original features $\mathop X\nolimits_{valid}^{original}$  with $N$ feature fields and $M_{valid}$ samples, candidate cross features $\mathop X\nolimits_{valid}^{cross}$ with $\varepsilon$ cross feature fields and $M_{valid}$ samples, labels $\mathop Y_{valid}$ with $M_{valid}$ samples, model weights $\mathop W$ of LR trained with both original and cross features on training set, lookup function $\rm{LOOKUP}()$ for one-dimensional embeddings in the sparse LR model, sigmoid function $\gamma()$ and AUC computing function $compute\_auc()$.
        \ENSURE final set of cross feature fields $\mathop S\nolimits^*$.
        \STATE $\mathop S\nolimits^* = \{\}$;
        \STATE $\mathop E\nolimits_{valid}^{original} = \rm{LOOKUP}\left(\mathop X\nolimits_{valid}^{original}\right)$;
        \STATE $\mathop E\nolimits_{valid}^{cross} = \rm{LOOKUP}\left(\mathop X\nolimits_{valid}^{cross}\right)$;
        \STATE $\mathop b\nolimits(-1) = \mathop E\nolimits_{valid}^{original} \mathop W[0:N]$;
        \STATE $AUC(-1) = compute\_auc(Y_{valid}, \gamma(\mathop b(-1)))$;
        \FOR{$i$ in $[0, \varepsilon)$}
            \FOR{$j$ in $[0, \varepsilon)$}
                \IF{$c_j$ not in $\mathop S\nolimits^*$}
                    \STATE $\mathop b(j) = \mathop b(-1) + \mathop E\nolimits_{valid}^{cross}[:,j]\mathop W[N+j]$;
                    \STATE $AUC(j) = compute\_auc(Y_{valid}, \gamma(\mathop b(j)))$;
                \ENDIF
            \ENDFOR
            \STATE $k = \mathop {{\rm{argmax}}}_j AUC(j)$;
            \IF{$AUC(k) > AUC(-1)$}
                \STATE $c_k \to \mathop S\nolimits^*$;
                \STATE $b(-1) = b(k)$;
                \STATE $AUC(-1) = AUC(k)$;
            \ELSE
                \STATE break;
            \ENDIF
        \ENDFOR
        \RETURN $\mathop S\nolimits^*$.
    \end{algorithmic}
\end{algorithm}

After obtaining the candidate set of cross feature fields $\mathop S = \{\mathop c_1, \mathop c_2, ...,$ $\mathop c_{\varepsilon}\}$, we can search for final useful cross feature fields.
The searching strategy in AutoCross \cite{yuanfei2019autocross} has been proven effective.
Facing a large candidate set, AutoCross trains a LR model for each candidate cross feature field, and select final cross feature fields based on their contribution measured on the validation set.
Obviously, there are too many LR models to train, and the efficiency is low.
In contrast, we have a compact and accurate candidate set, and are able to feed all candidate cross feature fields into a single LR model.
Thus, we do not need to involve the complex searching structure in \cite{yuanfei2019autocross}, and can simply select useful cross feature fields during the validation process.

To search for useful cross feature fields, we need to train a sparse LR model.
The input of the LR model consists of both original features and candidate cross features.
We use $\mathop S$ as the schema to process the training set to generate corresponding candidate cross features, which are denoted as $\mathop X\nolimits_{train}^{cross}$, with $\varepsilon$ cross feature fields and $M_{train}$ samples.
The original features of training set are denoted as $\mathop X\nolimits_{train}^{original}$, with $N$ feature fields and $M_{train}$ samples.
Accordingly, the number of all input feature fields for the LR model is $N+\varepsilon$.
As mentioned above, $\varepsilon = 2N$ or $\varepsilon = 3N$ is enough according to our experiments.
Therefore, the time cost of training the LR model with both original and cross features is in the same order of magnitude with directly training a LR model with only original features.

Based on the model weights $\mathop W$ in the trained LR model, we conduct the searching procedure for useful cross feature fields according to their contribution measured on the validation set.
Here, $\mathop X\nolimits_{valid}^{original}$ and $\mathop X\nolimits_{valid}^{cross}$ denote original features and candidate cross features on the validation set respectively.
Pseudocode of the searching procedure is presented in Alg. \ref{alg:searching}.
Moreover, the steps 7-12 in Alg. \ref{alg:searching} are paralleled with multi-threading implementation, where the measuring of each candidate cross feature field is conducted on one thread.
The main idea in Alg. \ref{alg:searching} is that, based on parameters in the trained LR model, we measure the contribution in AUC of each candidate cross feature field on the validation set, and iteratively select those have positive contribution as the final set $\mathop S\nolimits^*$ of cross feature fields.

\section{Experiments}\label{section:exp}

In this section, we empirically evaluate our proposed DNN2LR approach. We first describe settings of the experiments, then report and analyze the experimental results.
Thorough evaluations are conducted to answer the following research questions:
\begin{itemize}
	\item \textbf{RQ1} Can DNN2LR empower the simple LR model achieving good performances?
	\item \textbf{RQ2} How is the efficiency of DNN2LR for feature crossing, especially on datasets with large numbers of feature fields?
	\item \textbf{RQ3} How many cross feature fields can be automatically generated?
	\item \textbf{RQ4} How is the sensitivity of the DNN2LR method to hyper-parameters?
\end{itemize}

\subsection{Experimental Settings}

We evaluate the proposed DNN2LR method on $4$ public datasets, i.e., Employee\footnote{\scriptsize{\url{https://www.kaggle.com/c/amazon-employee-access-challenge/data}}}, Criteo\footnote{\scriptsize{\url{https://www.kaggle.com/c/criteo-display-ad-challenge/data}}}, Allstate\footnote{\scriptsize{\url{https://www.kaggle.com/c/allstate-claims-severity/data}}} and BNP\footnote{\scriptsize{\url{https://www.kaggle.com/c/bnp-paribas-cardif-claims-management/data}}}.
Meanwhile, we also conduct experiments on $2$ real-world datasets, i.e., RW1 and RW2, from our applications in finance, after anonymization and sanitization.
More details about these datasets can be found in Tab. \ref{tab:data}.
These datasets have one thing in common: DNN can outperform LR on these datasets, which means there are cross features to find.
According to the numbers of feature fields in Tab. \ref{tab:data}, we can conclude two kinds of datasets: narrow datasets and wide datasets.
On each dataset, we use $20\%$ of the training samples for validation.
Meanwhile, we transform the regression task in Allstate as binary classification tasks with settled thresholds, where labels less than $2550$ are set as negative, and positive otherwise.

In our experiments, we are going to verify whether we can obtain powerful and globally interpretable models.
Specifically, we are going to verify whether DNN2LR can empower the simple LR model to achieve better performances comparing with the complex DNN model, as well as other competitive feature crossing methods.
Accordingly, we give comparison among several types of methods: baselines, factorization-based methods, deep learning-based methods, tree-based methods, searching-based methods and interpretation-inspired methods.
We select some representative methods in each type and introduce them as follows.

\textbf{Baselines}: We incorporate sparse \textbf{LR} and sparse \textbf{DNN} as baselines, as introduced in Sec. \ref{sec:candidate}.
LR is a simple model, while DNN is a complex model.
For LR, we tune the learning rate in the range of $[0.001,1.0]$, and the l2 regularization in the range of $[0.0001,1.0]$.
For DNN, we set the dimensionality of feature embeddings as $10$, the learning rate as $0.001$, the l2 regularization as $0.0001$, and use Adam \cite{kingma2014adam} for optimization. We use the commonly-applied ReLu as activation function in hidden layers of DNN. The hidden components in deep layers of DNN are set as $[400, 200]$ on Criteo, and $[400, 100]$ on other datasets.

\textbf{Factorization-based}: We include \textbf{FM} \cite{rendle2010factorization}, \textbf{HOFM} \cite{blondel2016higher} and \textbf{AutoFM} \cite{liu2020autofis}.
FM is a successive method for modeling second-order feature interactions, and HOFM extends FM with higher-order feature interactions.
AutoFM directly learns the weight for each cross feature field, for both 2nd-order and higher-order.
We set the dimensionality of embeddings as $10$, the learning rate as $0.001$, the l2 regularization as $0.0001$.
We generate up to 4th-order cross features with HOFM.
The efficiency of AutoFM is low when datasets are wide and desired interaction order is high.
Thus, for efficiency, with AutoFM, we generate up to 4th-order cross features on Employee, 3rd-order cross features on Criteo and 2nd-order cross features on wide datasets.

\textbf{Deep learning-based}: We select \textbf{CrossNet} \cite{wang2017deep}, \textbf{CIN} \cite{lian2018xdeepfm} and \textbf{Self-Attention} \cite{song2018autoint,vaswani2017attention}.
CrossNet, CIN and Self-Attention are the crossing modules designed and applied in DCN \cite{wang2017deep}, xDeepFM \cite{lian2018xdeepfm} and AutoINT \cite{song2018autoint} respectively.
For those overlapping settings, we stay the same with above settings of DNN.
Moreover, for CrossNet, we have $3$ layers of feature crossing.
For CIN, we have $3$ layers of feature crossing, and the number of feature maps in the crossing module stays the same with the number of original feature fields.
For Self-Attention, we have $3$ layers of feature crossing, where the number of hidden units in self-Attention and the number of attention heads are set as $32$ and $2$ respectively.
These settings stay the same as in corresponding previous works, and $3$ layers of feature crossing means conducting up to $4$th-order crossing.

\begin{table}[!]
\centering
\caption{Summarization of the datasets.}\label{tab:data}
    \begin{tabular}{|c|cc|cc|c|}
    \hline
    \multirow{2}[2]{*}{dataset} & \multicolumn{2}{c|}{\#samples} & \multicolumn{2}{c|}{\#feature fields} & \multirow{2}[2]{*}{width} \\
          & training & testing & \#Num. & \#Cate. &  \\
    \hline
    Employee & 29,494 & 3,277 & 0     & 9     & narrow \\
    Criteo & 41.256M & 4.584M & 13    & 26    & narrow \\
    Allstate & 131,823 & 56,497 & 15    & 115   & wide \\
    BNP & 91,456 & 22,865 & 109   & 23    & wide \\
    RW1   & 233,123 & 58,282 & 178   & 15    & wide \\
    RW2   & 151,236 & 52,169 & 302   & 56    & wide \\
    \hline
    \end{tabular}%
\end{table}

\begin{table*}[htbp]
  \centering
  \caption{Performance comparison among different baselines and feature crossing methods in terms of AUC (\%). $*$ and $**$ denote statistically significant improvement, measured by t-test with p-value$<0.05$ and p-value$<0.01$ respectively, over the second best method on each dataset.}
  \label{tab:performance}
    \begin{tabular}{|c|cccccc|c|}
    \hline
    method & Employee & Criteo & Allstate & BNP   & RW1   & RW2   & average \\
    \hline
    LR    & 86.75  & 78.51  & 86.10  & 73.32  & 72.36  & 78.16  & 79.20  \\
    DNN   & 87.85  & 79.95  & 86.60  & 74.08  & 74.60  & 79.68  & 80.46  \\
    \hline
    FM    & 87.16  & 79.27  & 86.41  & 73.39  & 72.54  & 78.92  & 79.62  \\
    HOFM  & 87.28  & 79.72  & 86.48  & 73.46  & 72.78  & 79.21  & 79.82  \\
    AutoFM & 88.23  & 80.08  & 86.42  & 73.28  & 72.69  & 78.84  & 79.92  \\
    CrossNet & 87.52  & 79.57  & 86.38  & 73.43  & 73.94  & 79.07  & 79.99  \\
    CIN   & 88.12  & 80.05  & 86.49  & 73.64  & 74.06  & 79.38  & 80.29  \\
    Self-Attention & 88.33  & 79.89  & 86.53  & 73.97  & 74.39  & 79.51  & 80.44  \\
    GBDT+LR & 87.57  & 79.53  & 86.46  & 74.23  & 74.67  & 79.43  & 80.32  \\
    CMI+LR & 89.12  & 78.81  & 86.35  & 73.58  & 73.89  & 78.81  & 80.09  \\
    AutoCross+LR & 89.42  & 80.36  & 86.41  & 74.73  & 74.81  & 79.39  & 80.85  \\
    DNN2LR & $\ \ $\textbf{89.58}$^*$  & $\ \ \ \ $\textbf{80.48}$^{**}$  & $\ \ \ \ $\textbf{86.72}$^{**}$  & $\ \ \ \ $\textbf{75.48}$^{**}$  & $\ \ \ \ $\textbf{75.26}$^{**}$  & $\ \ \ \ $\textbf{79.91}$^{**}$  & $\ \ \ \ $\textbf{81.24}$^{**}$  \\
    \hline
    \end{tabular}%
\end{table*}%

\textbf{Tree-based}: We incorporate \textbf{GBDT+LR} \cite{he2014practical}, which generates cross features from GBDT.
Each tree in the GBDT model corresponds to a cross feature field in GBDT+LR.
Thus, we set the number of trees in GBDT as the same as the number of cross feature fields generated by DNN2LR on each dataset.

\textbf{Searching-based}: We incorporate \textbf{CMI} \cite{chapelle2015simple} and \textbf{AutoCross} \cite{yuanfei2019autocross}.
CMI searches for second-order cross feature fields based on conditional mutual information.
AutoCross is the state-of-the-art method for feature crossing, which can generate both second-order and higher-order cross feature fields.
The major setting of AutoCross is how many batches of data are divided for learning the contribution of each candidate cross feature field. As in \cite{yuanfei2019autocross}, on Employee, there are $2 \sum\nolimits_{i = 0}^{\left\lceil {\mathop {\log}\nolimits_2 t} \right\rceil  - 1} {\mathop 2\nolimits^i}$ batches of data, where $t$ is the size of candidate set in AutoCross. Meanwhile, on Criteo, there are $5 \sum\nolimits_{i = 0}^{\left\lceil {\mathop {\log}\nolimits_2 t} \right\rceil  - 1} {\mathop 2\nolimits^i}$ batches of data. For other datasets, i.e., Allstate, Movielens, RW1 and RW2, considering these datasets are relatively wide, we have $\sum\nolimits_{i = 0}^{\left\lceil {\mathop {\log }\nolimits_2 t} \right\rceil  - 1} {\mathop 2\nolimits^i }$ batches of data. The terminal condition of searching in AutoCross is set as, when newly added cross feature field leads to a performance degradation, the searching procedure stops.
Moreover, the cross features generated by CMI and AutoCross are fed into LR, and the corresponding performances are reported as \textbf{CMI+LR} and \textbf{AutoCross+LR} respectively.
The LR models used in these methods are set as above settings.


\textbf{Interpretation-inspired}: We include our proposed \textbf{DNN2LR} approach.
The quantile threshold $\eta$ for filtering feasible features in the interpretation inconsistency matrix is tuned in the range of $\{10\%,5\%,1\%,0.5\%,0.1\%\}$, and the size $\varepsilon$ of candidate set is tuned in the range of $\{ N, 2N, 3N, 4N, 5N \}$, where $N$ is the size of the original feature fields.
To verify the stability, we run DNN2LR $10$ times with different parameter initialization.
The LR models and DNN models used in DNN2LR are set as above settings.

Our experiments are conducted on a machine with Intel(R) Xeon(R) CPU (E5-26p30 v4 @ 2.20GHz, 24 cores) and 128G memory.
Moreover, we apply multi-granularity discretization \cite{yuanfei2019autocross} on numerical features.
The granularities in this method are set as $\mathop {\left\{ {\mathop {10}\nolimits^p} \right\}}\nolimits_{p = 1}^3$.

\begin{figure}
\centering
\includegraphics[width=0.48\textwidth]{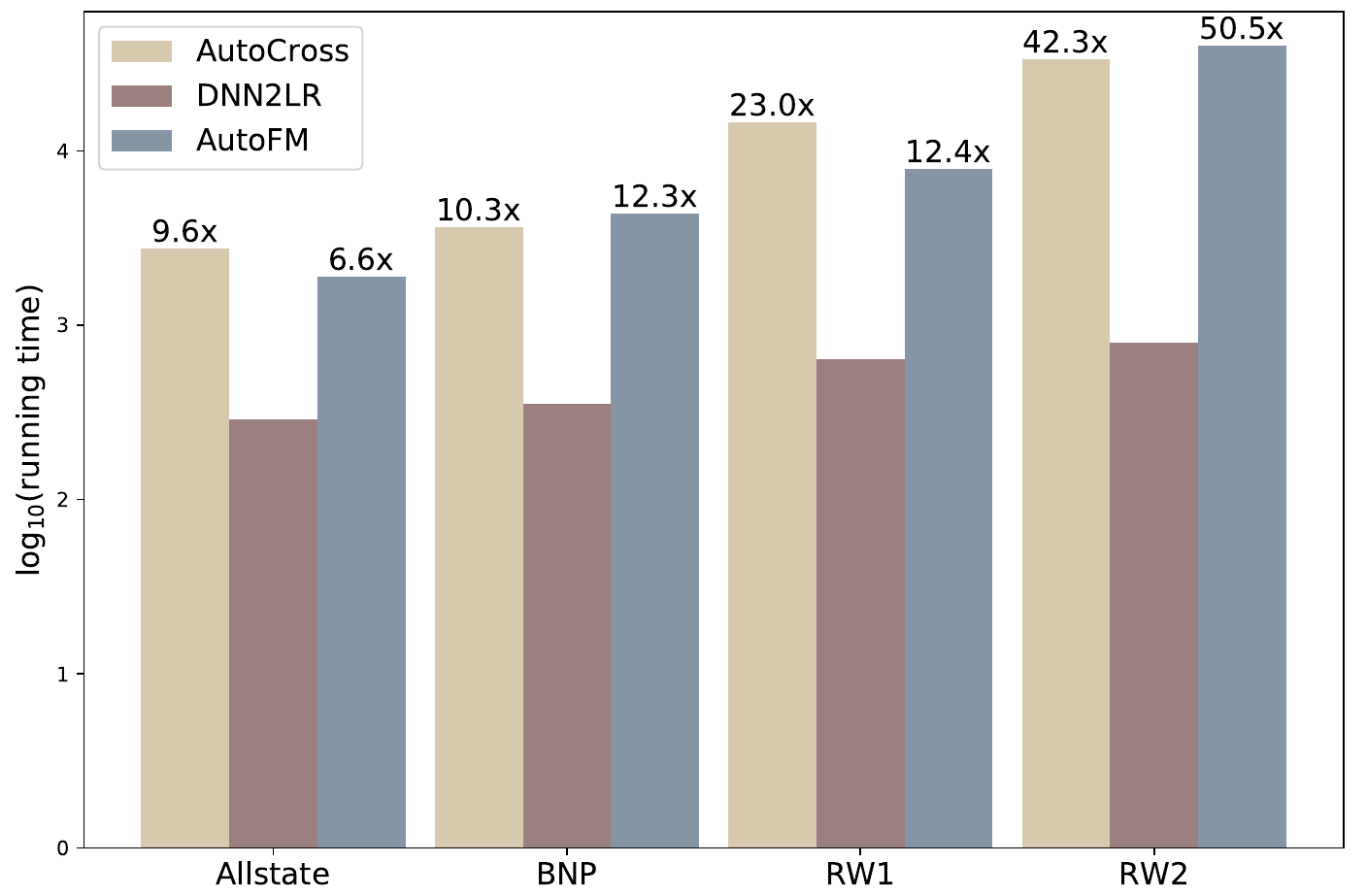}
\caption{Running time comparison among AutoCross, AutoFM and DNN2LR on wide datasets. We report running time in terms of seconds after the $log_{10}$ calculation. The ratios of running time of AutoCross and AutoFM to that of DNN2LR are respectively shown above the corresponding bars.}
\label{fig:speed}
\end{figure}

\subsection{Performance comparison (RQ1)}

Performance comparison is shown in Tab. \ref{tab:performance}.
On most datasets, different kinds of feature crossing methods can achieve somehow improvements on the performances of LR.
Among the three deep learning-based methods, CrossNet performs relatively poor. Meanwhile, CIN and Self-Attention achieve relatively good performances, but can not outperform DNN in average.
Nevertheless, as discussed in Sec. \ref{sec:related:FC}, these deep learning-based methods are not globally interpretable.
Similarly, GBDT+LR has near but lower performances compared with DNN.
CMI+LR, as a second-order searching-based method, has relatively poor overall performances.
This may caused by the fact that the CMI metric can not guarantee the effectiveness of generated cross features.
Meanwhile, AutoCross+LR outperforms DNN on Employee and Criteo, but performs relatively poor on wide datasets.
This is because that, on wide datasets with too many feature fields, data for training the field-wise LR model for evaluating a candidate cross feature field will be too little to produce reliable results. This causes the failure of AutoCross on wide datasets.
Clearly, DNN2LR is the only method that can constantly achieve powerful performances and outperform DNN and compared feature crossing methods on all datasets.
We have also run DNN2LR $10$ times with different parameter initialization and conducted significance test, which indicates that DNN2LR stably outperforms DNN, as well as other feature crossing methods, on both public and real-world datasets.

\begin{figure}
\centering
\includegraphics[width=0.48\textwidth]{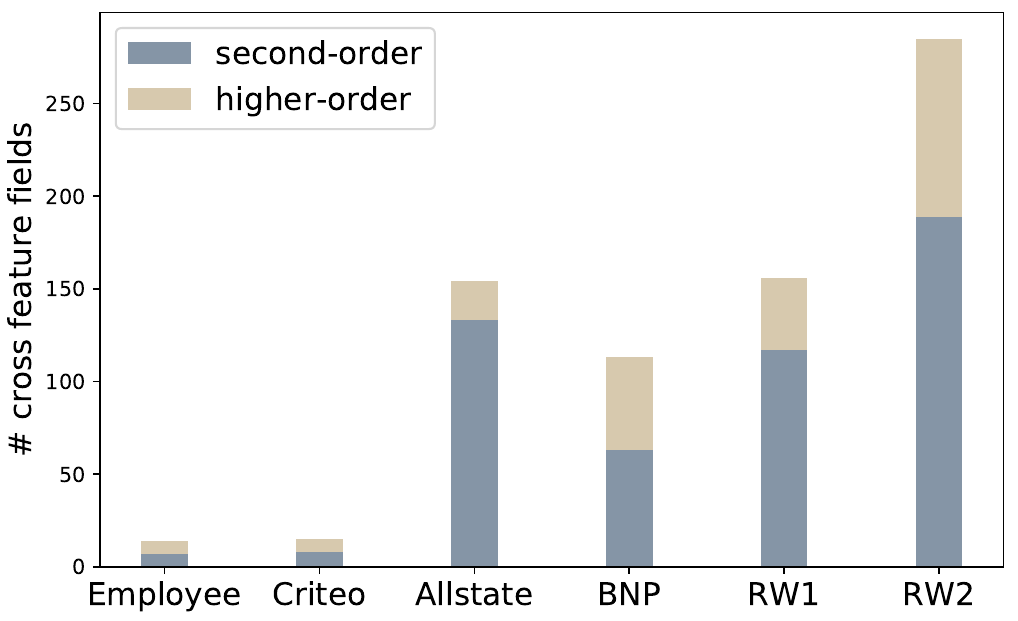}
\caption{The count of second-order and higher-order cross feature fields generated by DNN2LR on each dataset.}
\label{fig:count}
\end{figure}


\subsection{Running time for feature crossing (RQ2)}

For the efficiency comparison, we involve AutoCross and AutoFM.
Fig. \ref{fig:speed} shows the running time of AutoCross, AutoFM and DNN2LR for generating cross feature fields, which is illustrated after the $log_{10}$ calculation.
As comparing running time on wide datasets is much more valuable, we report running time on Allstate, BNP, RW1 and RW2 in Fig. \ref{fig:speed}.
We can clearly observe that, DNN2LR can perform much faster than AutoCross and AutoFM.
It is obvious that, compared with AutoCross and AutoFM, wider the dataset, larger DNN2LR's ratio of speedup.
That is to say, AutoCross and AutoFM have problem in efficiency on wide datasets.
As for DNN2LR, the whole procedure can be done via simply training a DNN model and a LR model, and the size of candidate set is only $3N$.
These results strongly illustrate the high efficiency of DNN2LR for feature crossing, especially on real-world datasets with large numbers of feature fields.

\subsection{Count of cross feature fields (RQ3)}

As shown in Fig. \ref{fig:count}, we illustrate the count of both second-order and higher-order cross feature fields generated by DNN2LR on each dataset.
Overall, we have more second-order cross feature fields than higher-order cross feature fields.
This may indicate that, second-order cross features are usually more important, and modeling higher-order cross features can bring further improvements.
Moreover, roughly speaking, more original feature fields we have, more cross feature fields in the final LR model we need.

\subsection{Sensitivity to hyper-parameters (RQ4)}

\begin{figure}
\centering
\includegraphics[width=0.48\textwidth]{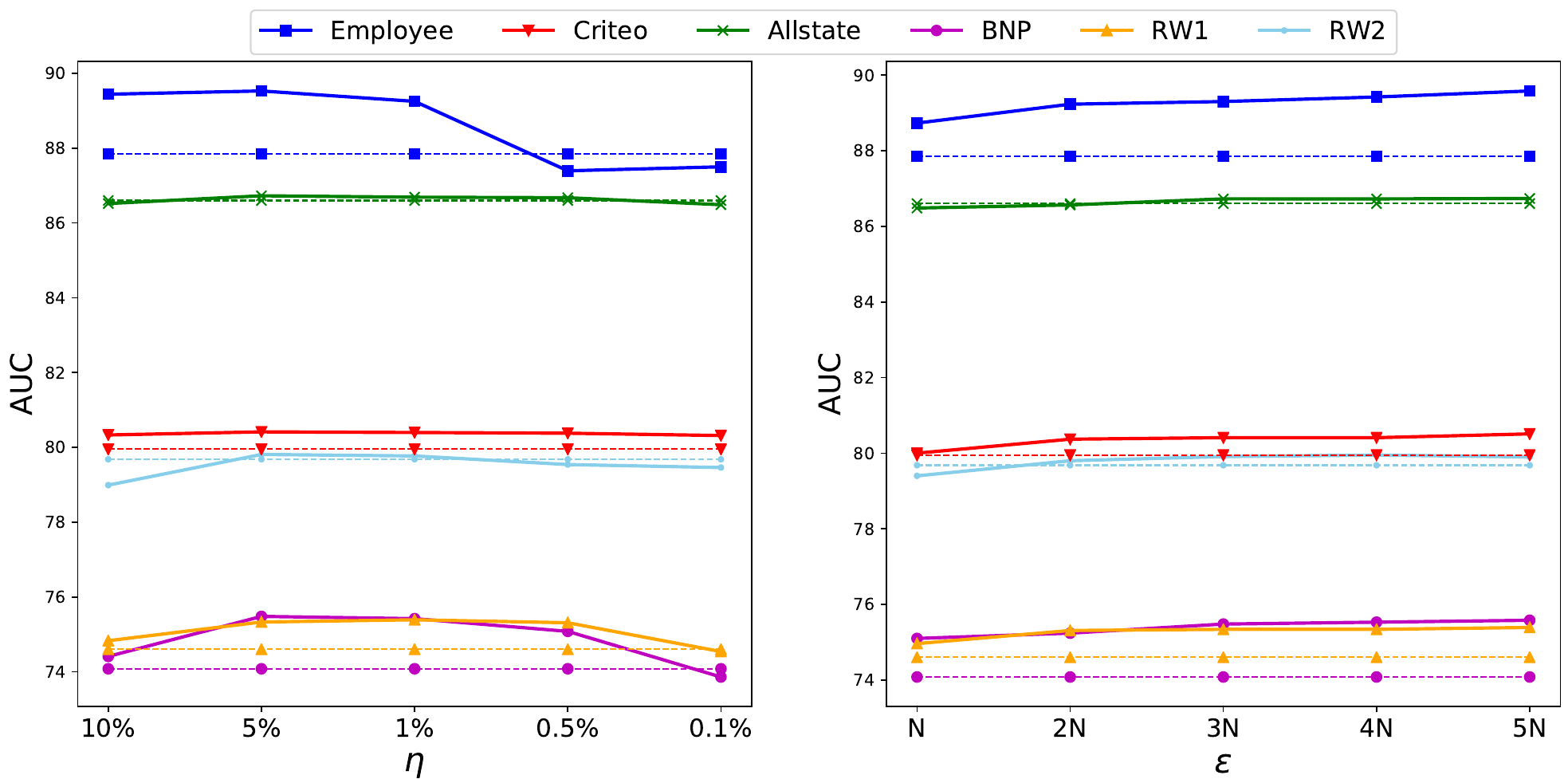}
\caption{Performances of DNN2LR with varying hyper-parameters: (1) the left part shows the impact of the quantile threshold $\eta$ for filtering feasible features in the interpretation inconsistency matrix; (2) the right part shows the impact of the size $\varepsilon$ of candidate set, where $N$ is the size of the original feature fields in each dataset. The dashed lines illustrate the performances of DNN on different datasets.}
\label{fig:params}
\end{figure}

As shown in Fig. \ref{fig:params}, we illustrate the sensitivity of DNN2LR to hyper-parameters.
First, we investigate the performances of DNN2LR with varying quantile thresholds $\eta$ for filtering feasible features in the interpretation inconsistency matrix.
According to the curves in Fig. \ref{fig:params}, the performances of DNN2LR are relatively stable, especially in the range of $\{5\%,1\%,0.5\%\}$.
The only exception is the Employee dataset, for it has only $9$ feature fields.
If we set $\eta$ too small, we will have too little feasible features for generating cross features on Employee.
Then, we investigate the performances of DNN2LR with varying sizes $\varepsilon$ of candidate set.
We can observe that, on datasets except Employee, the curves are stable when $\varepsilon \ge 2N$.
And when we have larger $\varepsilon$, the performances of DNN2LR slightly increase.
On Employee, with $\varepsilon = 5N$, we can have relative obvious performance improvement.
On each dataset, with only $\varepsilon = 2N$ or $\varepsilon = 3N$, DNN2LR can outperform the complex DNN model, where $N$ is the size of the original feature fields.
This makes the candidate size extremely small, comparing to the while set of possible cross feature fields.
According to our observations, we set $\eta = 5\%$ on all datasets, $\varepsilon = 5N$ on Employee, and $\varepsilon = 3N$ on other datasets as the optimal hyper-parameters in our experiments.

\section{Conclusions} \label{sec:conclusion}

In this paper, we observe that the local interpretations of DNN are usually inconsistent in different samples, and show that this is caused by feature interactions in the hidden layers of DNN.
Accordingly, we define the interpretation inconsistency in DNN, and propose a novel DNN2LR method.
DNN2LR can generate an accurate candidate set of cross feature fields, with extremely small amount compared to the whole set of second-order and higher-order cross feature fields in a dataset.
Based on the corresponding contribution in a LR model, useful cross feature fields can be directly ranked and selected from our compact candidate set.
The whole process of learning feature crossing can be done via simply training a DNN model and a LR model.
Extensive experiments have been conducted on four public datasets, as well as two real-world datasets.
Cross features generated by DNN2LR can empower a simple LR model achieving better performances comparing with the complex DNN model, as well as several state-of-the-art feature crossing methods, e.g., AutoCross and AutoFM.
The experiments also strongly verify the high efficiency of DNN2LR.
Especially on datasets with large numbers of feature fields, compared with AutoCross and AutoFM, DNN2LR can reduce the running time by $6\times$ to $50\times$.
In a word, with our proposed DNN2LR method, we can obtain powerful and globally interpretable models in an efficient way.

\ifCLASSOPTIONcaptionsoff
  \newpage
\fi



%
\balance
\bibliographystyle{IEEEtran}
\bibliography{DNN2LR}

\end{document}